\documentclass[10pt,twocolumn,letterpaper]{article}

\usepackage{arxiv}
\newtoggle{usetimes}
\newtoggle{itabs}
\toggletrue{usetimes}  
\usepackage{epsfig}
\usepackage{graphicx}
\usepackage{amsmath}
\usepackage{amssymb}
\usepackage{graphicx}
\usepackage{capt-of}
\usepackage{xcolor}
\usepackage{colortbl}
\usepackage{amsmath,amssymb,amsfonts,dsfont,pifont,bm,bbm,mathrsfs,mathtools,nicefrac}
\usepackage{algorithm,algpseudocode,listings}
\usepackage{booktabs,multirow,adjustbox,diagbox,threeparttable}
\usepackage{tabu}
\definecolor{citeblue}{RGB}{48,111,186}
\iftoggle{usetimes}{\usepackage{times}}{}
\usepackage{amsmath,amssymb,amsfonts,dsfont,pifont,bm,bbm,mathrsfs,mathtools,nicefrac}
\usepackage{algorithm,algpseudocode,listings}
\usepackage{booktabs,multirow,adjustbox,diagbox,threeparttable}
\usepackage[pagebackref=false,breaklinks=true,colorlinks=true,citecolor=citeblue,bookmarks=false]{hyperref}

\usepackage{cleveref}  

\crefname{section}{Sec.}{Secs.}
\Crefname{section}{Section}{Sections}
\crefname{table}{Tab.}{Tabs.}
\Crefname{table}{Table}{Tables}
\crefname{figure}{Fig.}{Figs.}
\Crefname{figure}{Figure}{Figures}
\crefname{equation}{Eq.}{Eqs.}
\Crefname{equation}{Equation}{Equations}
\crefname{theorem}{Thm.}{Thms.}
\Crefname{theorem}{Theorem}{Theorems}
\crefname{algorithm}{Alg.}{Algs.}
\Crefname{algorithm}{Algorithm}{Algorithms}
\usepackage{listings}
\definecolor{codegreen}{rgb}{0,0.6,0}
\definecolor{codegray}{rgb}{0.5,0.5,0.5}
\definecolor{codepurple}{rgb}{0.58,0,0.82}
\definecolor{backcolour}{rgb}{1.0,1.0,1.0}
\lstdefinestyle{mystyle}{
    backgroundcolor=\color{backcolour},
    commentstyle=\color{codegreen},
    keywordstyle=\color{magenta},
    numberstyle=\tiny\color{codegray},
    stringstyle=\color{codepurple},
    basicstyle=\ttfamily\scriptsize,
    breakatwhitespace=false,
    breaklines=true,
    captionpos=b,
    keepspaces=true,
    numbers=left,
    numbersep=5pt,
    showspaces=false,
    showstringspaces=false,
    showtabs=false,
    tabsize=2
}
\usepackage{array}
\definecolor{CQColor}{rgb}{0.0,0.0,1.0} 

\definecolor{CQRColor}{rgb}{1.0,0.0,1.0} 

\definecolor{CQXXYColor}{rgb}{1.0,0.0,0.0} 

\usepackage{bbding}
\usepackage{cite}
\definecolor{emphasizeColor}{HTML}{2755FF}

\definecolor{baseColor}{rgb}{0.75,0.05,0.1}

\definecolor{checkmarkColor}{rgb}{0.1,0.75,0.1}

\definecolor{avgcolor_bg}{RGB}{240,255,240}
\definecolor{avgcolor}{RGB}{84,139,84}
\definecolor{hidden_color}{RGB}{126,126,126}
\usepackage{colortbl}

\newcommand{\tocite}[1]{\textcolor{red}{[TO CITE]}}
\newcommand{\method}{ViM\xspace}
\newcommand{\supp}{\textit{Supplementary Material}\xspace}

\begin{document}

\title{ViM: Vision Middleware for Unified Downstream Transferring}
\author{Yutong Feng$^{1}$, Biao Gong$^{1}$, Jianwen Jiang$^{1}$, Yiliang Lv$^{1}$, Yujun Shen$^{2}$, Deli Zhao$^{1}$, Jingren Zhou$^{1}$ \\
{$^1$Alibaba Group}\ \ {$^2$Ant Group}\\
{\tt\small \{fengyutong.fyt, a.biao.gong, jianwen.alan, shenyujun0302, zhaodeli\}@gmail.com}\\
{\tt\small \{yiliang.lyl, jingren.zhou\}@alibaba-inc.com}\\ 
}

\maketitle

\begin{abstract}

\textit{Foundation models are pre-trained on massive data and transferred to downstream tasks via fine-tuning.
This work presents \textbf{Vision Middleware (\method)}, a new learning paradigm that targets unified transferring from a single foundation model to a variety of downstream tasks.
\method consists of a zoo of lightweight plug-in modules, each of which is independently learned on a midstream dataset with a shared frozen backbone.
Downstream tasks can then benefit from an adequate aggregation of the module zoo thanks to the rich knowledge inherited from midstream tasks.
There are three major advantages of such a design.
From the efficiency aspect, the upstream backbone can be trained only once and reused for all downstream tasks without tuning.
From the scalability aspect, we can easily append additional modules to \method with no influence on existing modules.
From the performance aspect, \method can include as many midstream tasks as possible, narrowing the task gap between upstream and downstream.
Considering these benefits, we believe that \method, which the community could maintain and develop together, would serve as a powerful tool to assist foundation models.}
\end{abstract}

\section{Introduction}\label{sec:intro}

\begin{figure}[t]
	\begin{center}
		\includegraphics[width=0.95\linewidth]{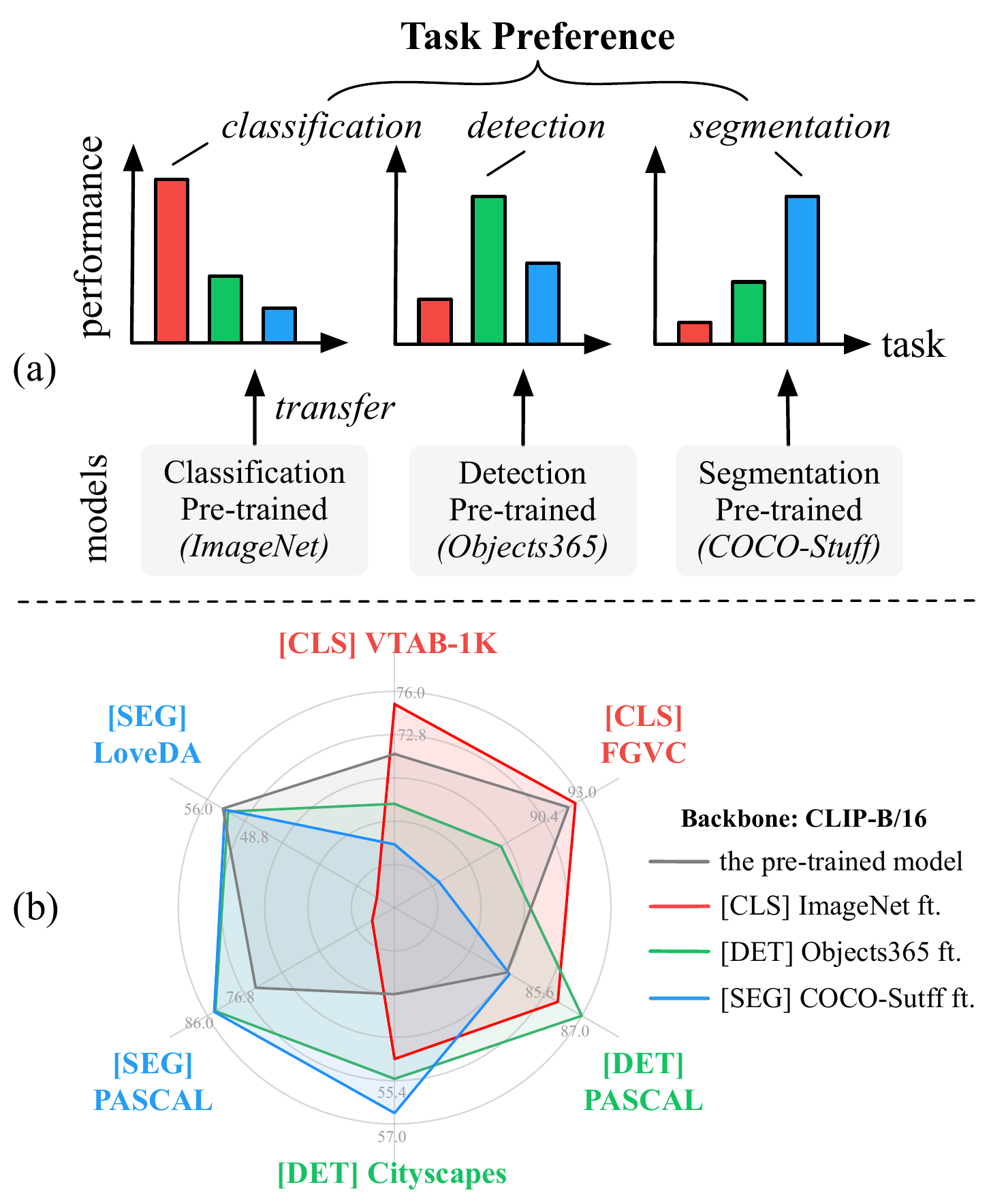}
	\end{center}
 \vspace{-16pt}
	\caption{\textbf{Task preference problem}: (a) \textit{Single-task pre-trained} models tend to perform better on tasks similar to their upstream. (b) \textit{Intermediate task fine-tuning} further tunes the pre-trained model on specific task, showing similar imbalanced performance.
}
\vspace{-12pt}
	\label{fig:intro}
\end{figure}

The \textit{pretrain-finetune} paradigm has served as a general framework across various vision tasks, where models are pre-trained on large-scale datasets and fine-tuned on downstream tasks~\cite{tan2018survey}.
Recent efforts have been attracted to build up a \textit{foundation model}~\cite{clip, coca, beit-3} with large architecture and massive pre-trained data (\textit{e.g.}, in the scale of billions).
Considering the trend of scaling up foundation models, 
it is costly to fine-tune model for different tasks separately,
and worthy of solving tasks with single model.
When directly transferred to downstream tasks, foundation models still suffer from the \textit{task-gap} problem.
Downstream tasks may require different targets with the upstream, thus could not fully leverage the pre-learned knowledge.
As shown in \Cref{fig:intro}, models are observed with \textit{preference} to those tasks similar to their encountered ones.
Task preference restricts unified transferring of single foundation model to multiple tasks with varying targets.

To bridge the task-gap, existing works address the problem on either upstream or downstream.
Upstream works propose multi-task pre-training to simultaneously learn various types of tasks~\cite{must, xlearner, gpv, ofa, unified-io}. 
However, when extending to a new pre-training task, these works require to re-formulate the task I/O and re-train the model together with existing tasks, which is complicated and time-consuming.
Downstream works introduce prompt \cite{ptuning-v2, vpt, noah} or adapter-based tuning~\cite{adapter, adaptformer, vit-adapter, clip-adapter, convpass} to adapt to  downstream tasks with additional parameters.
Since the downstream datasets for transferring are generally small (\textit{e.g.},~in the scale of thousands), the appended parameters might be insufficiently trained to master  newly-encountered tasks, usually resulting in unsatisfying performance~\cite{ppt, fewshot_survey}.


In this paper, we present a unified framework for supporting multiple downstream tasks with single foundation model. 
After the upstream pre-training, we introduce a collection of midstream tasks based on middle-scale datasets (\textit{e.g.}, in the scale of millions).
For each midstream task, a lightweight plug-in module is inserted into the pre-trained backbone, and individually optimized to learn the current task.
The foundation model is frozen without any parameter tuning in the above process.
Throughout this way, we collect the \textbf{Vision Middleware (\method)}, a module zoo containing knowledge from diverse midstream tasks based on the single foundation model.
To fully leverage the knowledge of ViM for downstream transferring, we develop practical strategies to adaptively aggregate ViM modules.
Modules correlated with the downstream task would be emphasized for better transferring.
ViM is expected to cope with the task-gap problem with sufficient supervision in midstream, and achieve balanced performance on multiple downstream tasks without preference.
During the experiments, we build up a ViM consisting of modules trained from 47 midstream tasks in 12 types, which can be grouped into global recognition, local recognition, vision and language understanding and self-supervised learning.
We then evaluate with the averaged transferring performance on 30 downstream tasks in 4 types.
The experimental results show satisfying performance of ViM with balanced improvements on multiple tasks.


ViM introduces a new paradigm of applying foundation models for unified transferring, which shows the following advantages:
(i) For the efficiency, the foundation model is maintained frozen after upstream pre-training, thus without any cost of storing different model parameters for various downstream tasks.
(ii) For the scalability, new ViM module can be individually trained and freely appended into the current ViM. The diverse middle-scale datasets from the community also enable us to easily expand ViM into great scale.
(iii) For the performance, ViM addresses the task-gap issue via including ViM modules of various midstream tasks, thus can achieve balanced performance for unified downstream transferring.
With the above advantages and verified performance, we would like to encourage the community to maintain a public ViM to which all researchers can contribute with their own well-trained ViM modules.
 
\begin{figure*}[t]
	\begin{center}
		\includegraphics[width=1.0\linewidth]{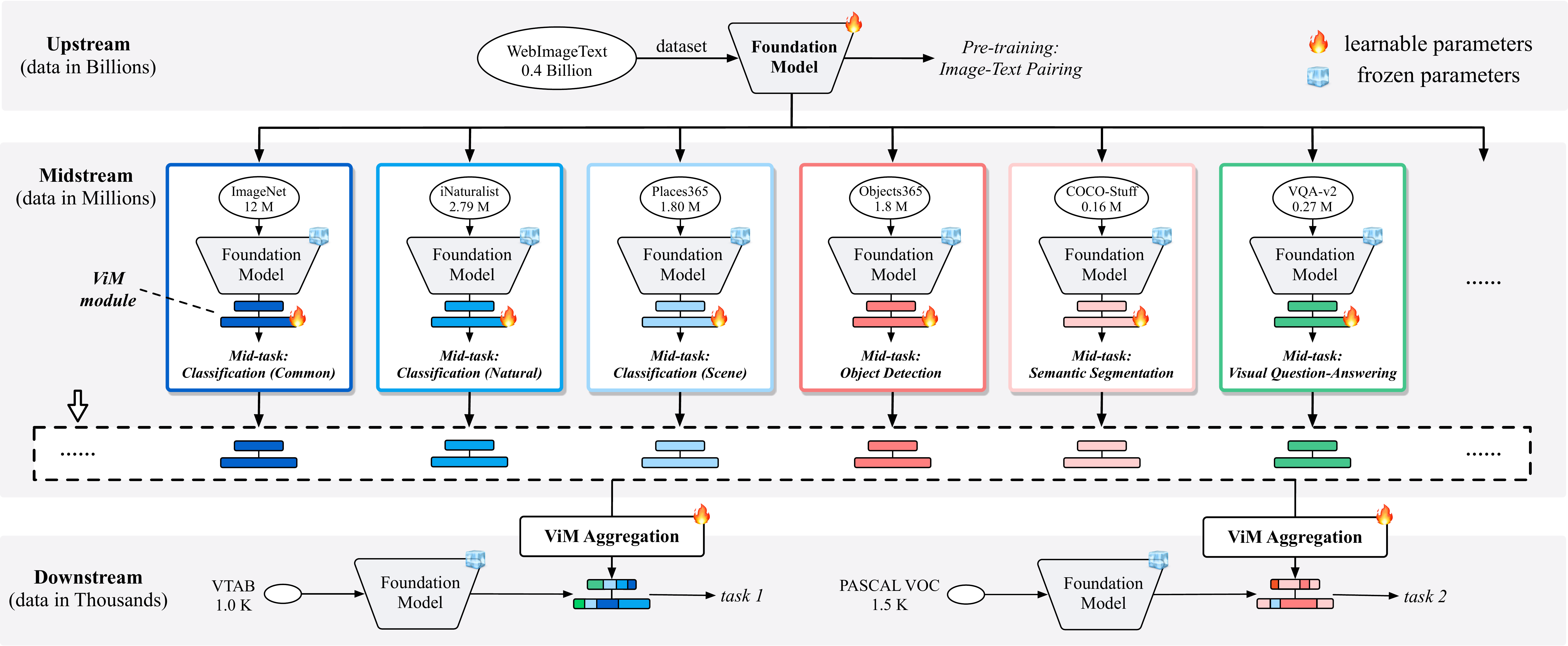}
	\end{center}
 \vspace{-14pt}
	\caption{%
	    \textbf{Vision Middleware} for \textit{unified} downstream transferring. The framework is in three stages based on varying sizes of datasets: (i) In the upstream, a foundation model is pre-trained leveraging a massive dataset (\textit{e.g.}, in billions). (ii) In the midstream, a sequence of different midstream tasks are introduced with middle-scale datasets (\textit{e.g.}, in millions). ViM is collected as a zoo of modules individually trained for each task, and the foundation model is frozen as the same backbone for all mid-tasks. (iii) In the downstream, ViM
	    aggregation adaptively gathers beneficial ViM modules  to work for each downstream dataset (\textit{e.g.}, in thousands).
	}
  \vspace{-10pt}
	\label{fig:pipeline}
\end{figure*}

\section{Related Work}\label{sec:related-work}
We review related works from perspectives of different stages for solving the task-gap problem.

\noindent\textbf{Upstream: multi-task pre-training.}
Multi-task pre-training~\cite{ruder2017overview} aims to train single model for supporting multiple tasks.  
%
\textit{Methods using individual heads:} Pioneer works study multi-task from single data source~\cite{overfeat,padnet,jtrl,mtinet,taskonomy,yu2020gradient}, where each sample contains multiple labels from different tasks, \textit{e.g.}, boundary, depth and segmentation maps. However, the single source scenario is restricted with expensive annotation and limited scenes.
Multi-source multi-task pre-training further combines different tasks from multiple datasets~\cite{ubernet, must,xlearner}, \textit{e.g.}, ImageNet~\cite{imagenet} and COCO~\cite{coco}. 
The multi-source scenario is much more free to benefit from diverse supervisions from varying tasks.
\textit{Methods using shared head:} More recently, studies on foundation models strive to unify the multi-task, multi-source and multi-modality pre-training~\cite{clip, coca, beit-3, gpv, ofa, unified-io}.
By converting different tasks into the same format of input and output (I/O), \textit{e.g.}, sequence to sequence~\cite{ofa,unified-io}, models can be pre-trained on them with both shared backbone and head.
Such a unification removes all the task-specific architectures, thus learns universal representation to master various tasks simultaneously.
\textit{Discussion:}
Multi-task pre-training involves various tasks to expand the task-coverage of the foundation model.
But it would be hard to continuously append new pre-training tasks, since all the tasks should be pre-trained together once again.

\noindent\textbf{Midstream: intermediate task fine-tuning.}
In the literature of natural language processing (NLP), it has been demonstrated that tuning pre-trained models on additional intermediate tasks can further boost downstream performance~\cite{stilts,pruksachatkun2020intermediate, DBLP:conf/emnlp/PothPRG21,chang2021rethinking}.
Experimental results indicate that not all intermediate tasks can benefit downstream tasks~\cite{DBLP:conf/emnlp/PothPRG21}, considering the task-similarity, level of semantics, domain distribution, \textit{etc.}
Recent application of vision foundation models also introduce intermediate fine-tuning before evaluating their transferrability, \textit{e.g.}, fine-tuning BEiT-3~\cite{beit-3} on Objects365~\cite{objects365} before transferring to COCO~\cite{coco}. 
The intermediate tasks are generally based on larger datasets, which enables fully supervision for understanding tasks.
Though with improved performance, since all the parameters of model have been fully adjusted for supporting single type of task, intermediate task fine-tuning can not be unified solution for multi-task transferring. 

\noindent\textbf{Downstream: parameter efficient tuning.}
Downstream solutions aim to adapt model with tuning a few parameters. 
\textit{Prompt-based tuning:} The prompt tuning is firstly introduced in NLP area to bridge the gap between upstream and downstream tasks~\cite{prompt_survery}.
Pioneer works define hand-crafted text templates as \textit{hard prompts} to convert downstream tasks into the format of pre-trained task~\cite{brown2020language,gao2020making},
which might be sub-optimal due to the man-made templates.
Continuous prompts are further proposed to assign additional trainable parameters as prompt for downstream tasks~\cite{ptuning-v2,lester2021power,liu2021gpt}.
VPT~\cite{vpt} firstly adapts prompt for vision models, where trainable prompts are appended as tokens in each layer.
\textit{Adapter-based tuning:} Adapters are lightweight modules inserted into the model for adapting tasks in NLP~\cite{adapter, lora}. Compared with prompt tuning, adapters can directly influence the model architecture for more generalized adaptation. 
For vision models, adapters have been widely applied on multiple tasks ~\cite{adaptformer, vit-adapter, clip-adapter, tip-adapter, convpass} with specific design for vision modalities.
\textit{Discussion:}
As a downstream solution, parameter efficient tuning is only supervised by the limited scale of downstream samples, which restricts the upper bound of its performance.

\section{Vision Middleware}
\label{sec:method}
We introduce Vision Middleware (ViM) to formulate a unified framework for supporting various vision tasks with single foundation model, as illustrated in \Cref{fig:pipeline}.
Following the framework, we present an implementation of building up and applying ViM, including the module architecture, midstream training and downstream aggregation.

\subsection{Towards Unified Downstream Transferring}
We start by reviewing the pretrain-finetune framework, where a foundation model is pre-trained on large-scale upstream dataset and fine-tuned on downstream tasks. 
Under such a framework, the downstream performance is correlated with the similarity between upstream and downstream tasks.
Disparate formats of tasks could lead to unsatisfying or even degraded downstream performance.
Such a task-gap restricts a pre-trained model to simultaneously support diverse downstream tasks.

To alleviate the task-gap problem for unified transferring with single foundation model, we introduce Vision Middleware (ViM) between the upstream and downstream stages, which factorizes varying abilities of model in advance to downstream transferring. 
ViM enables a new paradigm for assisting foundation model in the following three stages:

\begin{figure*}[t]
	\begin{center}
		\includegraphics[width=1.0\linewidth]{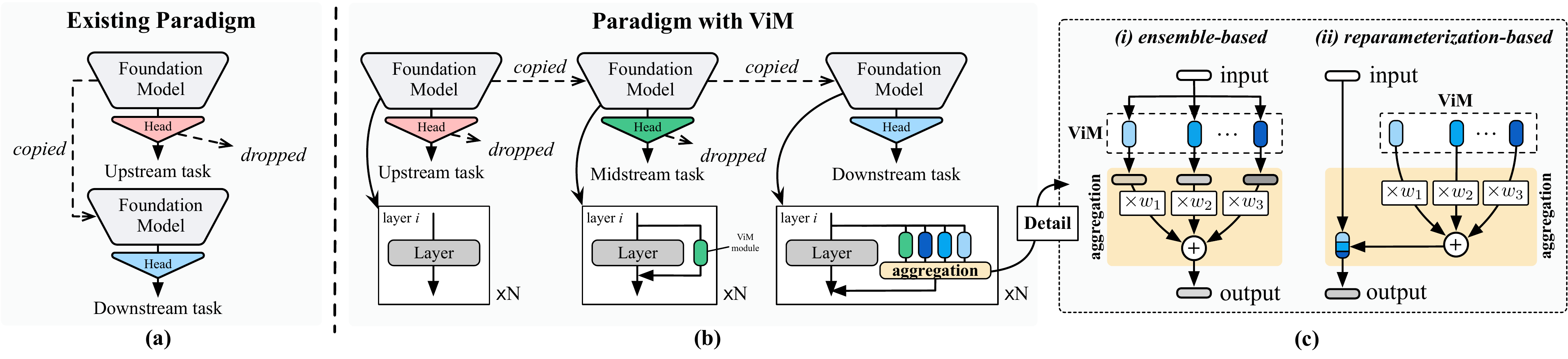}
	\end{center}
  \vspace{-18pt}
	\caption{(a) Existing transferring paradigm \textit{v.s.} (b) The detailed implementation of the \textbf{ViM transferring paradigm}. ViM modules are implemented as per-layer sub-modules and are aggregated in each layer. (c) Details of two \textbf{ViM aggregation} strategies.}
  \vspace{-8pt}
	\label{fig:method}
\end{figure*}

\noindent\textbf{Upstream:} A large-scale dataset with massive samples is leveraged to construct an upstream task $\mathcal{T}_{\text{up}}$, \textit{e.g.}, classification~\cite{he2019rethinking}, contrastive learning~\cite{clip, simclr, moco} and masked image modeling~\cite{beit,mae,simmim}.
Then a foundation model is supervised by task $\mathcal{T}_{\text{up}}$ with the optimized parameters $\Theta_{\text{up}}$.
Recent efforts concentrating on pre-training have been striving to expand the scale of both model architecture and pre-trained dataset, generating outstanding foundation models with fruitful knowledge.
It is noteworthy that the following processes can be used with arbitrary foundation model.

\noindent\textbf{Midstream:} In this stage, we introduce a set of midstream tasks, \textit{i.e.}, $\{\mathcal{T}_{\text{mid}}^1, \mathcal{T}_{\text{mid}}^2, ..., \mathcal{T}_{\text{mid}}^M\}$, based on different middle-scale datasets. 
The midstream tasks are expected to show diverse task formats, and allowed to be freely appended without limitation.
For each mid-task $\mathcal{T}_{\text{mid}}^i$, we assign a lightweight ViM module with parameters $\varphi^i_{\text{mid}}$, and combine it with the frozen foundation model for learning, which can be denoted as
$f([\Theta_{\text{up}}, \varphi_{\text{mid}}^i]) \rightarrow \mathcal{T}_{\text{mid}}^i$.
In the optimization process, only the parameters of the new ViM module are trainable for adapting to each mid-task. 
The original parameters of the foundation model are frozen to avoid damaging the pre-learned representation with task-specific shifting.
The ViM module should be in plug-in style towards the foundation model, but is not restricted for its design of architecture.
Besides, the module is expected to be relatively lightweight, \textit{i.e.}, $\Sigma_{i=1}^M \text{sizeof}(\varphi_{\text{mid}}^i) \ll \text{sizeof}(\Theta_{\text{up}}) $.
Such a lightweight characteristic enables us to collect ViM, \textit{i.e.}, a module zoo $\{\varphi_{\text{mid}}^i\}_{i=1}^{M}$, without large memory cost, while at the same time maintaining various abilities of the foundation model.

\noindent\textbf{Downstream:} Based on ViM inherited from the midstream, we aim to boost the performance on a wide range of downstream tasks. 
For each downstream task, ViM modules relevant to current task are expected to be picked up and aggregated together for transferring. 
Taking one task $\mathcal{T}_{\text{down}}$ as example, the
process can be denoted as
%
\begin{equation}
f([\Theta_{\text{up}}, ~\text{agg}(\varphi_{\text{mid}}^1, \varphi_{\text{mid}}^2, ..., \varphi_{\text{mid}}^M)]) \rightarrow \mathcal{T}_{\text{down}},
\end{equation}
where $\text{agg}(\cdot)$ is an aggregation function that adaptively incorporates different ViM modules for benefiting $\mathcal{T}_{\text{down}}$. 
It is noted that there may exist irrelevant modules in the zoo to current downstream task.
Since such cases are unknown for the midstream stage, the aggregation process is responsible to filter out these irrelevant modules, and maintain the performance on varying types of downstream tasks.

\noindent\textbf{Scalability:}
Thanks to the high extensibility, ViM could be maintained online by the community to continuously support different tasks.
Here we present a brief conceptual framework:
\textit{(i) Constructing ViM Module Zoo.} 
The zoo stores both the initial and developer uploaded ViM modules. 
Each module is specified with its training task, dataset and performance.
The task-specific head can be optionally stored together for direct usage.
\textit{(ii) Benchmarking and Grouping.} 
A benchmark with diverse tasks is set to evaluate ViM modules.
Then modules can be divided into groups based on their strengths, and serve as more lightweight ViMs for specific requirements.
\textit{(iii) Uploading Control.}
Any uploaded ViM module is firstly tested before appended into the zoo.
We would check the parameter sizes, avoid repeating with existing ones, and evaluate its transferability to filter out meaningless modules.
\textit{(iv) Usage.} 
Users can download pre-set ViM groups, or customize their own ViM zoo for specific target.
%
Corresponding algorithm of ViM aggregation  is also provided for transferring.



    



\subsection{Constructing and Applying ViM}
Following the framework, we present an implementation of how to build up, train and aggregate ViM. 
It should be noted that the following practice is not the only way of implementing the framework.
As an exploration, we demonstrate the potential of ViM for unified downstream transferring, and lay foundations for future works.

In this paper, we take the vision transformer (ViT)~\cite{vit} as the architecture of foundation model for presentation, which is commonly used in recent studies of pre-training ~\cite{clip,beit-3,coca}. 
ViT converts an image into a sequence of patch tokens, and encodes them using stacking of transformer layers with multi-head self-attention (MHSA)~\cite{transformer} and MLP modules.

\begin{table*}[t]
\caption{\textbf{Configurations} of the tasks and datasets for training ViM modules in the midstream. The numbers in brackets after each task type indicate the number of midstream tasks.
We only list the example datasets for brevity.
Additional task-specific architectures are assigned for adapting to each task, which are dropped after midstream training. Totally, we train on 12 types of tasks containing 47 mid-tasks, generating a zoo of diverse ViM modules with representation for various tasks.}
\label{tab:datasets}
\vspace{-6pt}
\small
\centering
\begin{tabular}{rlrl}
\toprule
\textbf{Task Type (\#mid-tasks)}                              & \textbf{Example Datasets}               & \textbf{Size} & \textbf{Task-specific Architecture}            \\
\hline
Image Classification (21) & ImageNet-21K~\cite{imagenet, in21k}, Places365~\cite{place365}, \textit{etc.}            & 26M  & Linear Classifier    \\
Object Detection (7)      & Objects365~\cite{objects365}, LogoDet-3K~\cite{logodet}, \textit{etc.}          & 2.3M & ViTDet~\cite{vitdet} \textit{w/} FastRCNN~\cite{fastrcnn} \\

Instance Segmentation (2) & LVIS~\cite{lvis}, COCO~\cite{coco}                        & 223K & ViTDet~\cite{vitdet} \textit{w/} MaskRCNN~\cite{maskrcnn} \\

Semantic Segmentation (4) & COCO-Stuff164K~\cite{cocostuff}, ADE20K~\cite{ade20k}, \textit{etc.}           & 240K & UperNet~\cite{upernet}            \\
                                          
Keypoints Estimation (1)                   & COCO-Keypoint~\cite{coco}                  & 57K  & ViTDet~\cite{vitdet} \textit{w/} KeypointRCNN~\cite{maskrcnn}              \\
Depth Estimation (2)      & NYU Depth V2~\cite{nyudepth}, KITTI (Eigen)~\cite{kitti, eigen2014depth}      &     48K  & Deconvolutions      \\
Visual Question Answering (2) & VQA-v2~\cite{vqa-v2}, GQA~\cite{gqa} &  265K & Concat. MLP with BERT~\cite{bert}\\
Referring Expression Compre. (3) & RefCOCO, RefCOCOg, RefCOCO+~\cite{refcoco} &  60K & MLM with PEVL text encoder~\cite{pevl}\\
Phrase Grounding (1) & Flickr30K~\cite{Flickr} & 32K & MLM with PEVL text encoder~\cite{pevl}\\
Visual Relationship Detection (1) & Visual Genome~\cite{VG} & 101K & MLM with PEVL text encoder~\cite{pevl}\\
Visual Commonsense Reasoning (1) & VCR~\cite{vcr} & 100K & MLM with PEVL text encoder~\cite{pevl}\\
Self-Supervised Learning (2) & ImageNet-1K~\cite{imagenet} & 1.3M & MoCo-v2~\cite{mocov2}, MAE~\cite{mae}\\
\hline
\textbf{TOTAL (47)} & & \textbf{32M} \\
\bottomrule
\end{tabular}
\vspace{-8pt}
\end{table*}

\noindent\textbf{Architecture of ViM module.} 
The ViM module can be implemented by any lightweight plug-in module to adapt the foundation model to specific tasks.
We adopt a recent adapter-based module, ConvPass~\cite{convpass},
for its simple architecture and stable performance on vision tasks.
%
%
As shown in \Cref{fig:method} (b), for each layer, an additional sequence of convolution layers are appended as a bypass.
It contains an $1\times1$ convolution squeezing the feature channel, a $3\times3$ convolution with the same input and output channels, and an $1\times1$ convolution recovering the channel.
The convolutions are concatenated with non-linear activation function, \textit{i.e.}, GELU~\cite{gelu}.
We append such a convolution sequence along with both the MHSA and MLP modules of all transformer layers.
With extremely small value of the squeezed channel (\textit{e.g.}, $1/96$ of the original dimension),  
the newly introduced parameters are only $0.4\%$ of the whole model.
Though with tiny size of learnable parameters, such a module has been proved with satisfying performance on task adaption.

\noindent\textbf{ViM Training.}
Based on the robust design of framework, each ViM module could be individually trained via different developing tools. Additional task-specific heads would be introduced depending on the current mid-task, \textit{e.g.}, linear classifier or RoI-Align~\cite{maskrcnn}.
These task-specific heads are possibly heavy, but would all be dropped after the midstream stage.
Only the trained ViM module is kept for memorizing the representation required by the mid-task.
Besides the mid-task supervised modules, we also append a \textit{zero} ViM module with parameters $\varphi^0_{\text{mid}}$, which is initialized to output all-zero values. 
The zero ViM module serves as a buffer to leave more space of downstream optimization.

\noindent\textbf{ViM Aggregation.}
Given the zoo of ViM with varying abilities of tasks, we aim to adaptively select and aggregate them for specific downstream task.
Different strategies of aggregation are explored based ensemble and reparameterization, as illustrated in \Cref{fig:method} (b).
The \textit{\textbf{ensemble}} is a common strategy to aggregate models from multiple pre-training sources.
Taking the input, we firstly forward it with different ViM modules, then aggregate their outputs.
To better activate relevant modules, we adopt the Mixture-of-Experts (MoE)~\cite{moe} that uses a routing module to pick up top-$k$ beneficial modules. 
Different from mapping the input to generate aggregation weights as in MoE, we directly optimize a vector of aggregation weights $\mathbf{w} \in \mathbb{R}^M$ for each task, and find it effective in experiments.
We activate the ViM modules with top-$k$ weight values and aggregate their outputs with weights after softmax.
By default, we activate all the modules into aggregation.
%
The \textit{\textbf{reparameterization}}-based strategy firstly aggregates the parameters of ViM modules into a single one, then only forward it once with better efficiency. We follow the same implementation of aggregation weights with the ensemble strategy. Since the aggregation is fixed after training, ViM could be employed as a single module during the inference.
%


\section{Experiments}
\label{sec:exp}

\subsection{Experimental Setup}

\noindent\textbf{Upstream Setup.} 
We utilize the CLIP~\cite{clip} pre-trained image encoder as the foundation model, due to its powerful generalization performance. 
CLIP is pre-trained with image-text contrastive learning on a web-scale dataset with 0.4 billion image-text pairs.
For the main experiments, we construct ViM based on the official ViT-B$/$16 model.

\begin{table*}[t]
\setlength\tabcolsep{0.9pt}
\caption{The \textbf{downstream transferring results} on 30 tasks in 4 groups of task types, \textit{i.e.}, image classification (\textbf{CLS}), object detection (\textbf{DET}), semantic segmentation (\textbf{SEG)} and depth estimation (\textbf{DEP}). 
Evaluation metrics for different tasks are shown in the brackets. 
The last column is highlighted with an averaged metric across all tasks.
The best results among methods with frozen CLIP-B/16 backbone are \textbf{bold} for each task.
\textbf{FB} indicates whether the method is based on frozen backbone, which is more flexible for multi-task transferring.
For the single-task pre-training, $\dag$ indicates the results are reported by INTERN~\cite{intern}, while $\ddag$ indicates the results are reported by MuST~\cite{must}.
}
\label{tab:exp_main}
\vspace{-6pt}
\small
\centering
\begin{tabular}{c@{\extracolsep{2pt}}c@{\extracolsep{6pt}}c@{\extracolsep{0pt}}c@{\extracolsep{2pt}}c@{\extracolsep{2pt}}c@{\extracolsep{4pt}}c@{\extracolsep{2pt}}ccccc@{\hspace{4pt}} >{\columncolor{avgcolor_bg}}c}
\toprule\\[-3.35ex]
 \multicolumn{4}{c}{\textbf{Method}}    &   \multicolumn{2}{c}{\textbf{CLS} (\footnotesize{acc.$\uparrow$)}} & \multicolumn{2}{c}{\textbf{DET} (\footnotesize{AP$_{50}\uparrow$)}} & \multicolumn{2}{c}{\textbf{SEG} \footnotesize{(mIoU$\uparrow$)}} & \multicolumn{2}{c}{\textbf{DEP} \footnotesize{(RMSE$\downarrow$~/~$\delta_1\uparrow$)}} &  \\[-0.1ex]
\cline{1-4}\cline{5-6}\cline{7-8}\cline{9-10}\cline{11-12}
\\[-4.1ex]
%
\multirow{5}{*}{\begin{tabular}[c]{@{}c@{}}\textbf{Upstream}\\ \textbf{Pre-training}\end{tabular}} & \multirow{5}{*}{\textbf{Midstream}} & \multirow{5}{*}{\begin{tabular}[c]{@{}c@{}}\textbf{Downstream}\\ \textbf{Tuning}\end{tabular}} &  \footnotesize \multirow{5}{*}{(\textbf{FB})} & \rotatebox{270}{VTAB (19)~}               & \rotatebox{270}{FGVC (5)}              & \rotatebox{270}{PASCAL}                & \rotatebox{270}{Cityscapes}                & \rotatebox{270}{PASCAL}                  & \rotatebox{270}{LoveDA}                 & \rotatebox{270}{NYUv2}                 & \rotatebox{270}{KITTI}        &  \multirow{4}{*}{\large {\textbf{Avg.}}}    \\[-0.4ex]
\midrule

\rowcolor[gray]{0.95} \multicolumn{13}{l}{\textit{\textbf{~single-task pre-training}}} \\
ImageNet~\cite{imagenet} Cls. $\dag$ & - & Linear & \ding{51} & \color{hidden_color}{-} & \color{hidden_color}{-} & \color{hidden_color}{82.7} & \color{hidden_color}{-} & \color{hidden_color}{67.8} & \color{hidden_color}{-} &  \color{hidden_color}{0.43~/~~~-~~~~} & \color{hidden_color}{3.06~/~~~-~~~~} & \color{hidden_color}{-}\\
CLIP-R50x16~\cite{clip} $\dag$ & - & Linear & \ding{51}& \color{hidden_color}{-} & \color{hidden_color}{-} & \color{hidden_color}{83.6} & \color{hidden_color}{-} & \color{hidden_color}{68.7} & \color{hidden_color}{-} &  \color{hidden_color}{0.39~/~~~-~~~~} & \color{hidden_color}{2.83~/~~~-~~~~} & \color{hidden_color}{-}\\
MoCo-v2~\cite{mocov2} $\dag$ & - & Linear & \ding{51}& \color{hidden_color}{-} & \color{hidden_color}{-} & \color{hidden_color}{79.1} & \color{hidden_color}{-} & \color{hidden_color}{66.9} & \color{hidden_color}{-} &  \color{hidden_color}{0.43~/~~~-~~~~} & \color{hidden_color}{3.09~/~~~-~~~~} & \color{hidden_color}{-}\\
JFT~\cite{jft} Cls. $\ddag$ & - & Fully & \ding{55} & \color{hidden_color}{-} & \color{hidden_color}{-} & \color{hidden_color}{85.2} & \color{hidden_color}{-} & \color{hidden_color}{80.4} & \color{hidden_color}{-} &  \color{hidden_color}{~~~-~~~/~86.5} & \color{hidden_color}{-} & \color{hidden_color}{-} \\
SimCLR (w. JFT)~\cite{simclr} $\ddag$ & - & Fully & \ding{55} & \color{hidden_color}{-} & \color{hidden_color}{-} & \color{hidden_color}{84.1} & \color{hidden_color}{-} & \color{hidden_color}{74.9} & \color{hidden_color}{-} &  \color{hidden_color}{~~~-~~~/~84.8} & \color{hidden_color}{-} & \color{hidden_color}{-} \\
\rowcolor[gray]{0.95} \multicolumn{13}{l}{\textit{\textbf{~multi-task pre-training}}} \\
MuST (w. JFT)~\cite{must}       &    -                &       Fully     & \ding{55}     &       \color{hidden_color}{-}                     &       \color{hidden_color}{-}                &         \color{hidden_color}{87.9}                &        \color{hidden_color}{-}                   &           \color{hidden_color}{82.9}              &           \color{hidden_color}{-}             &            \color{hidden_color}{~~~-~~~/~89.5}           &        \color{hidden_color}{-}        & \color{hidden_color}{-}  \\
X-Learner$_{\text{R152}}$~\cite{xlearner}     &   -     & Fully      & \ding{55}            &          \color{hidden_color}{-}                  &      \color{hidden_color}{-}                 &           \color{hidden_color}{88.6}              &        \color{hidden_color}{-}                   &          \color{hidden_color}{82.6}               &             \color{hidden_color}{-}           &            \color{hidden_color}{~~~-~~~/~91.3}           &            \color{hidden_color}{-}        & \color{hidden_color}{-}   \\
Unified-IO$_{\text{LARGE}}$~\cite{unified-io}        & - &      Fully     & \ding{55}      &           \color{hidden_color}{-}                 &           \color{hidden_color}{-}            &          \color{hidden_color}{-}               &             \color{hidden_color}{-}              &         \color{hidden_color}{-}               &            \color{hidden_color}{-}            &              \color{hidden_color}{0.40~/~~~-~~~~}         &             \color{hidden_color}{-}       & \color{hidden_color}{-}   \\
INTERN~\cite{intern} \scriptsize{(1B param.)}       &   -           &  Linear  & \ding{51}                  &            \color{hidden_color}{-}                &            \color{hidden_color}{-}          &           \color{hidden_color}{90.7}              &            \color{hidden_color}{-}               &           \color{hidden_color}{78.7}              &        \color{hidden_color}{-}               &        \color{hidden_color}{0.32~/~~~-~~~~}               &         \color{hidden_color}{2.55~/~~~-~~~}       & \color{hidden_color}{-}       \\
\midrule
\rowcolor[gray]{0.95} \multicolumn{13}{l}{\textit{\textbf{~intermediate fine-tuning~/~parameter efficient tuning}}}          \\
\multirow{8}{*}{CLIP-B/16~\cite{clip}}           &     -      &  Fully    & \ding{55}  & \color{hidden_color}{71.4}                      & \color{hidden_color}{92.1}                 &           \color{hidden_color}{84.2}              &             \color{hidden_color}{49.8}              & \color{hidden_color}{74.1}                    & \color{hidden_color}{53.0}                   & \color{hidden_color}{0.37~/~90.0}                 &          \color{hidden_color}{2.42~/~96.0}     & \color{hidden_color}{76.3}      \\
& ImageNet21K ft. & ConvPass~\cite{convpass} & \ding{51} &            75.1                &           \textbf{92.6}           &          86.1               &          53.4                 &            45.5             &         23.4               &          0.44~/~84.8             &           2.75~/~94.5         & 69.4  \\
& Objects365 ft.     &      ConvPass~\cite{convpass}        & \ding{51}  &     67.7             &            87.4                &        87.0               &          54.5               &            84.0               &           52.0              &       0.42~/~86.3                 &              2.64~/~94.4           & 76.7       \\
& COCO-Stuff ft. &    ConvPass~\cite{convpass}     & \ding{51} &     64.7               &              83.1              &             84.3          &             56.4            &              84.3             &           52.6              &               0.44~/~85.2               &          2.82~/~93.9       & 75.8     \\
& - & Linear    & \ding{51}          &              61.7              &        81.7               &            48.5             &              37.6             &            78.6             &          49.0              &           0.61~/~70.1            &          3.49~/~88.4       & 64.4     \\
&-  & VPT~\cite{vpt}            & \ding{51}                       &            71.3                &       85.3                &          -               &            -               &            82.8             &           49.6             &           0.50~/~80.0            &          3.05~/~92.1      & -      \\
&-  & Adapter~\cite{adapter}       & \ding{51}                   &             73.1               &          90.1             &           84.7              &           54.4                &         84.5                &           51.8             &            0.43~/~85.4           &           2.70~/~94.3       & 77.3    \\
&-  & ConvPass~\cite{convpass}        & \ding{51}                          &              73.1              &           90.2            &             84.9            &            55.1               &          84.4               &            53.0            &            0.41~/~86.9           &          2.52~/~95.1       & 77.8     \\
\midrule
\rowcolor[gray]{0.95} \multicolumn{13}{l}{\textit{\textbf{~Ours: Vision Middleware (ViM)}}}                                                                                                                                                                                                   \\
\multirow{2}{*}{CLIP-B/16~\cite{clip}}  & +ViM                & ViM-agg (rep.) & \ding{51} &             74.8               &          90.2             &              85.9           &           55.6                &             85.0            &        52.6                &          0.39~/~88.7             &  2.46~/~95.6 &       78.6            \\
& +ViM                &       ViM-agg (ens.)         & \ding{51}            &              \textbf{75.3}              &          91.7             &      \textbf{87.2}                   &             \textbf{56.9}              &         \textbf{86.0}                &         \textbf{54.1}     & \textbf{0.38}~/~\textbf{89.0}          &             \textbf{2.38~}/~\textbf{96.3}           &\textbf{79.6}           \\
\bottomrule
\end{tabular}
\vspace{-5pt}
\end{table*}

\noindent\textbf{Midstream Setup.} 
As shown in \Cref{tab:datasets}, we incorporate 12 types of midstream tasks with totally 47 sub-tasks for training ViM. 
The tasks can grouped into global recognition (\textit{e.g.}, image classification), local recognition (\textit{e.g.}, detection and segmentation), vision-language (\textit{e.g.}, VQA and visual grounding) and self-supervised learning (\textit{e.g.} masked image modeling).
Additional task-specific architectures are assigned for the varying targets on each task.
%
%
Taking the zero ViM module into account, we collect a ViM with size of 48.
More details of the midstream configurations and training results are in the \supp.

\noindent\textbf{Downstream Setup.} We evaluate the transferrability of the proposed framework on the following downstream tasks:
\textit{\textbf{(i) Classification:}} We leverage two common benchmarks for fine-grained classification, \textit{i.e.}, the VTAB-1k~\cite{vtab} and FGVC benchmark. 
VTAB-1k contains $19$ datasets divided into Natural, Specialized and Structured groups with different contents. 
FGVC consists of 5 datasets including the CUB-200-2011~\cite{cub200}, NABirds~\cite{nabirds}, Oxford Flowers~\cite{flowers}, Stanford Dogs~\cite{stanforddogs} and Stanford Cars~\cite{stanfordcars}.
Linear classifier is appended as the task head for classification learning.
\textit{\textbf{(ii) Object Detection:}} We evaluate with the PASCAL~\cite{voc} dataset (07+12) for common scene, and the Cityscapes~\cite{cityscapes} dataset for traffic scene. 
For the task head of detection, ViTDet~\cite{vitdet} is used with FastRCNN~\cite{fastrcnn} for PASCAL and MaskRCNN~\cite{maskrcnn} for Cityscapes.
\textit{\textbf{(iii) Semantic Segmentation:}} We evaluate with the PASCAL~\cite{voc} (2012 set) dataset for common scene, and the LoveDA~\cite{loveda} dataset for remote sensing.
We reshape the feature maps from layer $4, 6, 8, 12$ into varying resolutions with up/down sampling as in~\cite{beit}, then feed them into a semantic FPN module~\cite{fpn} for segmentation map prediction.
\textit{\textbf{(iv) Depth Estimation:}} We evaluate with the NYU Depth V2~\cite{nyudepth} and KITTI~\cite{kitti} (with~\cite{eigen2014depth} split) datasets for indoor and outdoor scenes, respectively. We append deconvolution and up-sampling layers to predict the depth map.
\textit{\textbf{Summary:}} We evaluate with 30 tasks in 4 types. Details of datasets and training configurations are in \supp.

\noindent\textbf{Compared Methods.} (i) We firstly compare the results of single or multiple task \textit{pre-training} as reported. Since they are based on various backbones and tuning methods, we list the results for reference.
(ii) For the \textit{intermediate task fine-tuning}, we respectively fine-tune the model with the largest datasets of 3 tasks in the midstream, \textit{i.e.}, classification on ImageNet-21K~\cite{imagenet, in21k}, detection on Objects365~\cite{objects365} and segmentation COCO-Stuff164K~\cite{cocostuff}.
The fine-tuned models are transferred to downstream with SoTA tuning method.
(iii) For the \textit{parameter efficient tuning}, we compare with recent SoTA tuning methods for vision~\cite{vpt, adapter, convpass}.
%
%
(iv) The two ViM aggregation methods  are denoted as ViM-agg (rep. or ens.).
We optimize the aggregation weights, activated ViM modules and task head for downstream task learning.

\subsection{Downstream Transferring Results}
\noindent\textbf{Analysis on Main Results.} \Cref{tab:exp_main} shows the transferring results on all downstream tasks.
To investigate the unified transferability, we compute an averaged metric across different tasks (taking the $\delta_1$ for depth estimation).
Compared with single-task pre-training, multi-task pre-training models generally show balanced performance on different tasks. 
As we have discussed, they require to include all the task types in pre-training stage, and lack extensibility to more tasks.
The intermediate fine-tuned models achieve great improvement on similar tasks, but also decrease dramatically on the remaining tasks. 
For instance, the ImageNet-21K fine-tuned model performs well on VTAB classification ($+3.7\%$ accuracy than fully tuning), but shows $-28.6\%$ mIoU on PASCAL segmentation.
Such an imbalanced performance is also illustrated in ~\Cref{fig:intro} (b).
For the parameter efficient tuning, \textit{i.e.}, VPT~\cite{vpt}, adapter~\cite{adapter} and ConvPass~\cite{convpass}, they show general improvement to linear probing on all tasks.
However, such improvements are restricted by the limited downstream data for mastering new tasks.

When introducing ViM in the midstream and conducting ViM aggregation for downstream, we achieve the highest performance on the averaged metric and almost all tasks. 
It is noteworthy that we conduct fair comparison on all methods with the same CLIP-B/16 backbone.
The ensemble-based aggregation shows better improvements with $+3.3\%$ than fully tuning without touching the backbone.
The reparameterization-based aggregation, though performs slightly worse than ensemble, is also more effective than compared methods, and shows better efficiency on the computation cost during the inference.
With a frozen backbone, ViM is also competitive with multi-task pre-training models using fully tuning or larger backbone.
On some dense prediction tasks, \textit{e.g.}, PASCAL detection and NYUv2, MuST~\cite{must} and X-Learner~\cite{xlearner} achieve better results, which is attributed to the natural gap between their convolutional backbone and our plain ViT backbone.

\noindent\textbf{ViM with different backbone.} 
The design of our framework is decoupled with backbone.
To further demonstrate the effectiveness of ViM, we also construct a ViM based on the CLIP ViT-L/14 backbone, with the 21 classification mid-tasks. 
We further transfer it to downstream VTAB classification. 
\Cref{tab:change_backbone} shows that ViM maintains improvement with larger backbone and higher baseline results. 

\begin{table}[t]
\caption{Results of changing the \textbf{backbone} to CLIP-L/14.}
\vspace{-6pt}
\small
\centering
\begin{tabular}{cc@{\extracolsep{6pt}}c@{\extracolsep{6pt}}c@{\extracolsep{6pt}}c@{\extracolsep{6pt}}c@{\extracolsep{6pt}}}
\toprule
\textbf{Backbone} & \textbf{Method} & \textbf{VTAB} & \footnotesize -Natural &\footnotesize -Special. & \footnotesize -Struct. \\
\midrule
\multirow{2}{*}{CLIP-L/14} & ConvPass & 76.81 & 84.41 & 87.78 & 58.25\\
 & \textbf{ViM} & \textbf{78.07} & \textbf{85.02} & \textbf{88.13} & \textbf{61.06} \\
\bottomrule
\end{tabular}
\label{tab:change_backbone}
\vspace{-2pt}
\end{table}
\begin{table}[t]
\centering
\caption{Results of \textbf{further transferring to midstream tasks}. ``Mid.'' indicates training single ViM module in the midstream.}
\footnotesize
\vspace{-6pt}
\begin{tabular}{c@{\extracolsep{2pt}}c@{\extracolsep{2pt}}c@{\extracolsep{2pt}}c@{\extracolsep{2pt}}c@{\extracolsep{2pt}}c}
\toprule
\textbf{Method}         & \footnotesize \textbf{IN1K}~\cite{imagenet} & \footnotesize \textbf{iNat18}~\cite{inat18}  &\footnotesize \textbf{THDogs}~\cite{thdogs} &\footnotesize \textbf{Logo}~\cite{logo2k} &\footnotesize \textbf{FoodX}~\cite{foox251} \\
[-0.4ex]
\midrule
\rowcolor{gray!20}
Fully   & 87.89             & 81.93              & 92.05     & 92.39       & 84.90        \\
Mid. & 86.60             & 78.40               & 89.94    &     91.37     & 83.37        \\
\textbf{ViM}  & 87.74             & 81.18             & 91.37     &        92.31           & 84.80       \\
\bottomrule
\end{tabular}
\label{tab:mid_trans}
\vspace{-8pt}
\end{table}

\noindent\textbf{Transferring ViM to midstream tasks.}
We also apply the same transferring process back to the midstream tasks for training ViM.
\Cref{tab:mid_trans} shows the results on some midstream datasets based on the above ViM with CLIP-L/14.
Since the backbone is frozen, the performance of single ViM module trained in midstream is worse  than fully tuning.
However, with further transferring ViM to midstream tasks, we successfully compensate the performance degradation without touching the backbone, even comparable to fully tuning, demonstrating the effectiveness of our paradigm.

\noindent\textbf{Transferring ViM to unseen tasks.}
For specific downstream task, the ViM modules trained on similar mid-tasks play important roles.
To further investigate their importance, we construct experiments to transfer ViM to unseen tasks, \textit{e.g.}, removing the classification ViM modules and transferring to classification.
As shown in \Cref{tab:unseen_tasks}, removing similar midstream modules indeed decreases the corresponding results, but still achieves better performance than compared methods.
The results suggest that ViM modules on similar mid-tasks are \textit{important but not the only beneficial modules} for downstream transferring.
The detailed study on contribution of modules is in \Cref{sec:vis}.

\noindent\textbf{Transferring with single ViM module.}
The transferability of single ViM module is studied in \Cref{tab:single_vim}.
We observe similar unbalanced performance as the intermediate task fine-tuning, where modules generally perform better only on tasks correlated with their encountered ones.

\begin{table}[t]
\caption{Results of transferring to \textbf{unseen tasks} in the midstream.}
\vspace{-6pt}
\small
\centering
\begin{tabular}{cc@{\extracolsep{8pt}}c@{\extracolsep{8pt}}c@{\extracolsep{9pt}}c}
\toprule
\textbf{ViM (size)  }    &   \begin{tabular}[c]{@{}c@{}}\textbf{CLS}\\ VTAB\end{tabular} & \begin{tabular}[c]{@{}c@{}}\textbf{DET}\\ PASCAL\end{tabular} & \begin{tabular}[c]{@{}c@{}}\textbf{SEG}\\ PASCAL\end{tabular} & \begin{tabular}[c]{@{}c@{}}\textbf{DEP}\\ KITTI\end{tabular} \\
\midrule
\rowcolor{gray!20} \textit{ConvPass (ref.)} & 73.1 & 84.9 & 84.4 & 2.52 / 95.1\\
\textit{w/o} CLS  (37)     &          75.0                                          &               -                                       &                   -                                &                     -                                  \\
\textit{w/o} DET (39)     &                      -                              &           86.6                                           &                               -                    &                           -                            \\
\textit{w/o} SEG (42)      &                            -                       &                 -                                     &             85.9                                      &                        -                               \\
\textit{w/o} DEP (46)     &                      -                         &              -                                        &                         -                          &                          2.51 / 95.5                             \\
\rowcolor{gray!20} whole set (48)   &               75.3                                     &                  87.2                                    &                       86.0                            &                2.38 / 96.3                   \\
\bottomrule
\end{tabular}
\label{tab:unseen_tasks}
\vspace{-6pt}
\end{table}

\begin{table}[t]
\caption{Directly transferring \textbf{single ViM module}, which is named as [Task]-[Dataset]. The best value of single module is \underline{underlined}.}
\vspace{-6pt}
\small
\centering
\begin{tabular}{c@{\extracolsep{4pt}}c@{\extracolsep{4pt}}c@{\extracolsep{2pt}}c@{\extracolsep{2pt}}c@{\extracolsep{3pt}}c@{\extracolsep{2pt}}c}
\toprule
\multirow{2}{*}{\textbf{Method}} & \multirow{2}{*}{\textbf{AVG}} & \multicolumn{3}{c}{\textbf{CLS} (VTAB)}                                       & \multicolumn{2}{c}{\textbf{SEG}}                     \\
\cline{3-5}\cline{6-7}
                        &                      & \footnotesize -Natural              & \footnotesize -Special.             & \footnotesize -Struct.              & \footnotesize PASCAL               & \footnotesize LoveDA               \\
\midrule
CLS-IN1K                & 71.8                 & \underline{78.9}                 & \underline{86.1}                 & 57.1                 & 84.1                 & 53.0                 \\
CLS-Places              & 71.5                 & 77.2                 & 85.9                 & \underline{58.7}        & 83.3                 & 52.3                 \\
DET-LVIS                & 70.8                 & 74.4                 & 84.3                 & 56.7                 & 83.9                 & \underline{53.4}                 \\
SEG-ADE20K              & 70.4                 & 74.1                 & 84.2                 & 56.2                 & \underline{85.2}                 & 52.2                 \\
\rowcolor{gray!20} \textbf{ViM (zoo)}               &  \textbf{73.2}  &  \textbf{79.9} & \textbf{87.2} & \textbf{58.9} & \textbf{86.0} & \textbf{54.1}          \\
\bottomrule
\end{tabular}
\label{tab:single_vim}
\vspace{-10pt}
\end{table}

\subsection{Ablation Studies}
We conduct ablation studies within the process of ViM aggregation, the results of which are expected to be helpful for further improved implementation of our framework.

\begin{figure}[t]
	\begin{center}
		\includegraphics[width=0.9\linewidth]{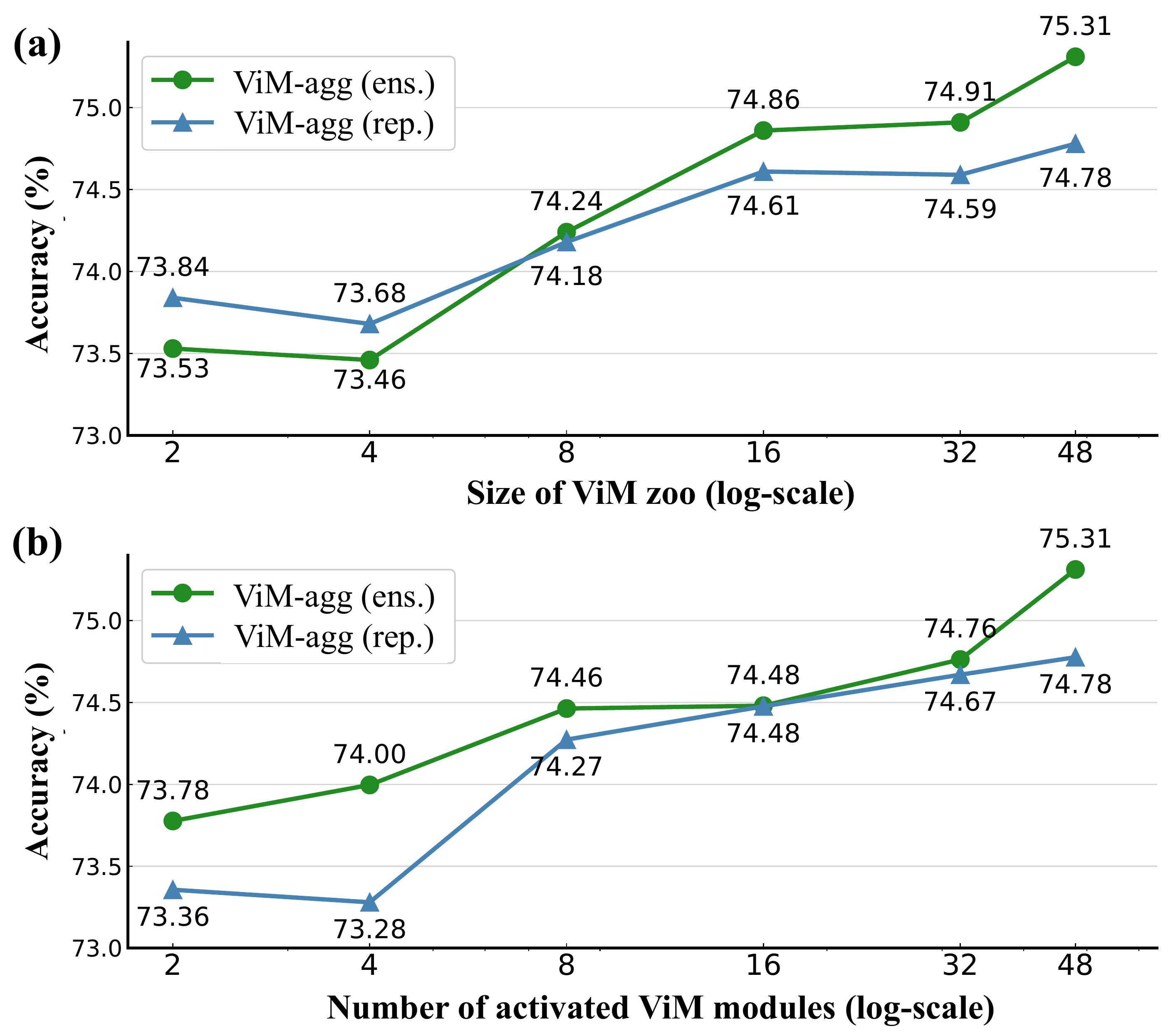}
	\end{center}
	\vspace{-20pt}
	\caption{Ablation study on (a) the \textbf{ViM zoo size} and (b) the \textbf{number of activated ViM modules}, conducted on VTAB classification with two aggregation strategies.}
	\label{fig:ablation}
    \vspace{-5pt}
\end{figure}

\noindent\textbf{On the size of ViM.} 
To verify the significance of expanding ViM with more modules, we construct ViM in varying sizes via randomly adding modules into the zoo.
 \Cref{fig:ablation} (a) shows the results.
The downstream performance is observed to be positively related to the zoo size of ViM, which encourages us to continuously append new ViM modules from diverse mid-tasks for better transferring.

\noindent\textbf{On the number of activated ViM modules.}
By default, all the ViM modules are activated.
We study the influence of activating different number of modules in \Cref{fig:ablation} (b).
%
Both showing increasing performance with more activate modules, the ensemble-based aggregation consistently outperforms reparameterization-based strategy.
%
%
Therefore, when training with more available memory, it is suggested to increase the number of activated modules and firstly consider the ensemble-based strategy.

\noindent\textbf{On the modeling of aggregation weights.} 
The aggregation weights are set as trainable vectors by default. 
We also explore to generate weights using a projection of the input embeddings, as in~\cite{moe, zoo_tuning}.
\Cref{tab:agg_weights} shows the results of projection using linear layer or MLP.
Though with more trainable parameters, it is hard to learn to project the input into proper aggregation weights, especially with a linear projection. 
Therefore, the trainable vector is a simple yet effective way of modeling aggregation weights.

\noindent\textbf{On the parameters of ViM modules.}
To demonstrate the improvement of ViM is not introduced by more parameters, we conduct experiment with \textit{empty modules}, which are set as the initial state of ViM modules without midstream training.
As shown in \cref{tab:empty_vim}, directly introducing more parameters without our midstream training would not result in similar improvement as ViM, and may cause even worse performance for transferring.
%

\begin{table}[t]
\caption{Ablation study on \textbf{modeling of aggregation weights} with ViM-agg (rep.).  ``\#params'' indicates the modeling parameters.}
\vspace{-6pt}
\small
\centering
\begin{tabular}{l@{\extracolsep{6pt}}c@{\extracolsep{4pt}}c@{\extracolsep{4pt}}c@{\extracolsep{4pt}}c@{\extracolsep{6pt}}c}
\toprule
\textbf{Agg. weights} & \textbf{VTAB} & \footnotesize -Natural &\footnotesize -Special. & \footnotesize -Struct. & \textbf{\#params} \\
\midrule
proj. (linear) & 18.49 & 3.99 & 37.51 & 13.97 & 0.1758 M\\
proj. (MLP) & 73.04 & 78.95	 & 85.99 & 54.19 & 0.4281 M\\
\rowcolor{gray!20} vector & 74.78 & 79.14 & 86.21 & 58.98 & 0.0002 M\\
\bottomrule
\end{tabular}
\label{tab:agg_weights}
\vspace{-4pt}
\end{table}

\begin{table}[t]
\caption{Ablation study on \textbf{empty ViM modules}.}
\vspace{-6pt}
\small
\centering
\begin{tabular}{l@{\extracolsep{6pt}}c@{\extracolsep{4pt}}c@{\extracolsep{4pt}}c@{\extracolsep{4pt}}c@{\extracolsep{6pt}}r}
\toprule
\textbf{Method} & \textbf{VTAB} & \footnotesize -Natural &\footnotesize -Special. & \footnotesize -Struct. & \textbf{\#params} \\
\midrule
empty ($\times 48$) & 71.77 & 78.55	 & 84.40 & 52.36 & 14.98 M\\
\rowcolor{gray!20} ViM ($\times 48$) & 75.31 & 80.18 & 86.80 & 58.95 & 14.98 M\\
\bottomrule
\end{tabular}
\label{tab:empty_vim}
\vspace{-12pt}
\end{table}

\begin{figure}[t]
	\begin{center}
		\includegraphics[width=1.0\linewidth]{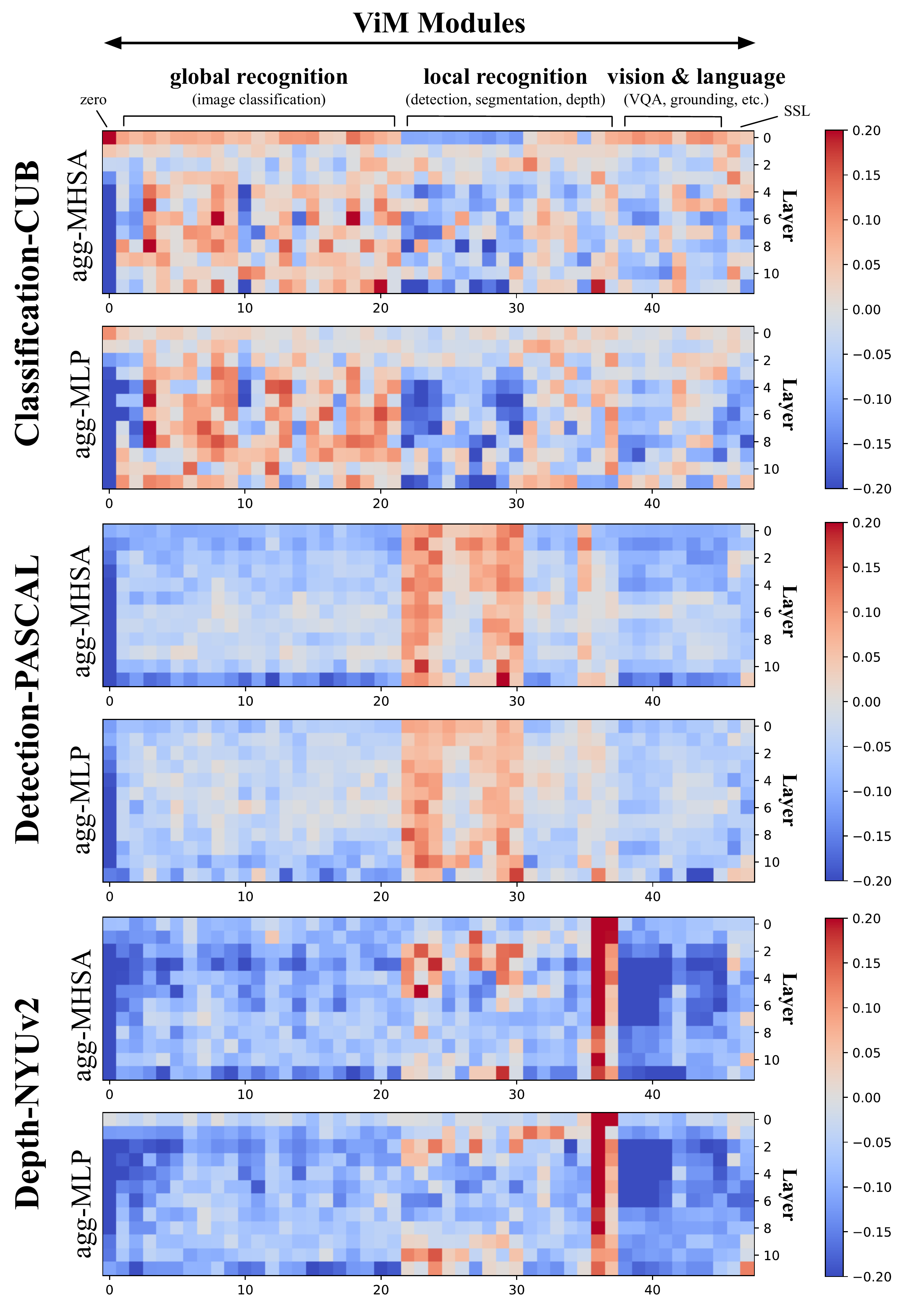}
	\end{center}
	\vspace{-20pt}
	\caption{\textbf{Visualization of the aggregation weights} on different downstream tasks. The activation of ViM modules along with the MHSA and MLP inside each layer are visualized, respectively.}
	\label{fig:agg_weights}
    \vspace{-6pt}
\end{figure}

\subsection{Visualization Analysis}
\label{sec:vis}
For further understanding the influence of different ViM modules in the aggregation process, we visualize the aggregation weights of all layers in \Cref{fig:agg_weights}.
We label the ViM modules trained with different groups of mid-tasks.
It is observed that ViM modules with similar mid-tasks to the current task are emphasized in general, \textit{e.g.}, modules with local recognition abilities show higher weights on downstream detection task.
Nevertheless, modules with different mid-tasks could also contribute from their own perspectives and knowledge to further improve the performance.
We also observe that ViM modules are aggregated with different weights among layers, suggesting that different modules are found beneficial in varying depth of the network, and work in a per-layer cooperation manner.

\section{Conclusion}\label{sec:conclusion}
In this paper, we present the Vision Middleware (ViM) in a unified framework to support multiple vision tasks with single foundation model.
ViM contains a zoo of lightweight modules, each of which is trained individually on a midstream task for factorizing task-specific knowledge with shared backbone.
For downstream transferring, the ViM modules are adaptively aggregated to boost the performance on various downstream tasks.
Experimental results 
indicate that ViM achieves balanced and improved performance.
The efficiency, extensibility and effectiveness of ViM encourage the community to maintain a public ViM, where modules are contributed from different researchers to facilitate the application of foundation models.

{\small
\bibliographystyle{abbrv}
\bibliography{ref}

\begin{thebibliography}{10}

\bibitem{beit}
H.~Bao, L.~Dong, S.~Piao, and F.~Wei.
\newblock Beit: Bert pre-training of image transformers.
\newblock In {\em Int. Conf. Learn. Represent.}, 2021.

\bibitem{brown2020language}
T.~Brown, B.~Mann, N.~Ryder, M.~Subbiah, J.~D. Kaplan, P.~Dhariwal,
  A.~Neelakantan, P.~Shyam, G.~Sastry, A.~Askell, et~al.
\newblock Language models are few-shot learners.
\newblock {\em Adv. Neural Inform. Process. Syst.}, pages 1877--1901, 2020.

\bibitem{cocostuff}
H.~Caesar, J.~R.~R. Uijlings, and V.~Ferrari.
\newblock Coco-stuff: Thing and stuff classes in context.
\newblock In {\em IEEE Conf. Comput. Vis. Pattern Recog.}, pages 1209--1218,
  2018.

\bibitem{chang2021rethinking}
T.-Y. Chang and C.-J. Lu.
\newblock Rethinking why intermediate-task fine-tuning works.
\newblock {\em arXiv preprint arXiv:2108.11696}, 2021.

\bibitem{adaptformer}
S.~Chen, C.~Ge, Z.~Tong, J.~Wang, Y.~Song, J.~Wang, and P.~Luo.
\newblock Adaptformer: Adapting vision transformers for scalable visual
  recognition.
\newblock {\em arXiv preprint arXiv:2205.13535}, 2022.

\bibitem{simclr}
T.~Chen, S.~Kornblith, M.~Norouzi, and G.~E. Hinton.
\newblock A simple framework for contrastive learning of visual
  representations.
\newblock In {\em Int. Conf. Mach. Learn.}, pages 1597--1607, 2020.

\bibitem{mocov2}
X.~Chen, H.~Fan, R.~Girshick, and K.~He.
\newblock Improved baselines with momentum contrastive learning.
\newblock {\em arXiv preprint arXiv:2003.04297}, 2020.

\bibitem{vit-adapter}
Z.~Chen, Y.~Duan, W.~Wang, J.~He, T.~Lu, J.~Dai, and Y.~Qiao.
\newblock Vision transformer adapter for dense predictions.
\newblock {\em arXiv preprint arXiv:2205.08534}, 2022.

\bibitem{cityscapes}
M.~Cordts, M.~Omran, S.~Ramos, T.~Rehfeld, M.~Enzweiler, R.~Benenson,
  U.~Franke, S.~Roth, and B.~Schiele.
\newblock The cityscapes dataset for semantic urban scene understanding.
\newblock In {\em IEEE Conf. Comput. Vis. Pattern Recog.}, pages 3213--3223,
  2016.

\bibitem{imagenet}
J.~Deng, W.~Dong, R.~Socher, L.-J. Li, K.~Li, and L.~Fei-Fei.
\newblock Imagenet: A large-scale hierarchical image database.
\newblock In {\em IEEE Conf. Comput. Vis. Pattern Recog.}, pages 248--255,
  2009.

\bibitem{bert}
J.~Devlin, M.~Chang, K.~Lee, and K.~Toutanova.
\newblock {BERT:} pre-training of deep bidirectional transformers for language
  understanding.
\newblock In {\em Proceedings of the 2019 Conference of the North American
  Chapter of the Association for Computational Linguistics: Human Language
  Technologies}, pages 4171--4186. Association for Computational Linguistics,
  2019.

\bibitem{voc}
Z.~Dong, K.~Xu, Y.~Yang, H.~Bao, W.~Xu, and R.~W.~H. Lau.
\newblock Location-aware single image reflection removal.
\newblock In {\em Int. Conf. Comput. Vis.}, pages 4997--5006, 2021.

\bibitem{vit}
A.~Dosovitskiy, L.~Beyer, A.~Kolesnikov, D.~Weissenborn, X.~Zhai,
  T.~Unterthiner, M.~Dehghani, M.~Minderer, G.~Heigold, S.~Gelly, et~al.
\newblock An image is worth 16x16 words: Transformers for image recognition at
  scale.
\newblock In {\em Int. Conf. Learn. Represent.}, 2020.

\bibitem{eigen2014depth}
D.~Eigen, C.~Puhrsch, and R.~Fergus.
\newblock Depth map prediction from a single image using a multi-scale deep
  network.
\newblock {\em Adv. Neural Inform. Process. Syst.}, 2014.

\bibitem{clip-adapter}
P.~Gao, S.~Geng, R.~Zhang, T.~Ma, R.~Fang, Y.~Zhang, H.~Li, and Y.~Qiao.
\newblock Clip-adapter: Better vision-language models with feature adapters.
\newblock {\em arXiv preprint arXiv:2110.04544}, 2021.

\bibitem{gao2020making}
T.~Gao, A.~Fisch, and D.~Chen.
\newblock Making pre-trained language models better few-shot learners.
\newblock {\em arXiv preprint arXiv:2012.15723}, 2020.

\bibitem{kitti}
A.~Geiger, P.~Lenz, C.~Stiller, and R.~Urtasun.
\newblock Vision meets robotics: The kitti dataset.
\newblock {\em The International Journal of Robotics Research}, pages
  1231--1237, 2013.

\bibitem{must}
G.~Ghiasi, B.~Zoph, E.~D. Cubuk, Q.~V. Le, and T.-Y. Lin.
\newblock Multi-task self-training for learning general representations.
\newblock In {\em Int. Conf. Comput. Vis.}, pages 8856--8865, 2021.

\bibitem{fastrcnn}
R.~Girshick.
\newblock Fast r-cnn.
\newblock In {\em Int. Conf. Comput. Vis.}, pages 1440--1448, 2015.

\bibitem{vqa-v2}
Y.~Goyal, T.~Khot, D.~Summers{-}Stay, D.~Batra, and D.~Parikh.
\newblock Making the {V} in {VQA} matter: Elevating the role of image
  understanding in visual question answering.
\newblock In {\em IEEE Conf. Comput. Vis. Pattern Recog.}, pages 6325--6334,
  2017.

\bibitem{ppt}
Y.~Gu, X.~Han, Z.~Liu, and M.~Huang.
\newblock Ppt: Pre-trained prompt tuning for few-shot learning.
\newblock {\em arXiv preprint arXiv:2109.04332}, 2021.

\bibitem{lvis}
A.~Gupta, P.~Doll{\'{a}}r, and R.~B. Girshick.
\newblock {LVIS:} {A} dataset for large vocabulary instance segmentation.
\newblock In {\em IEEE Conf. Comput. Vis. Pattern Recog.}, pages 5356--5364,
  2019.

\bibitem{gpv}
T.~Gupta, A.~Kamath, A.~Kembhavi, and D.~Hoiem.
\newblock Towards general purpose vision systems: An end-to-end task-agnostic
  vision-language architecture.
\newblock In {\em IEEE Conf. Comput. Vis. Pattern Recog.}, pages 16399--16409,
  2022.

\bibitem{mae}
K.~He, X.~Chen, S.~Xie, Y.~Li, P.~Doll{\'a}r, and R.~Girshick.
\newblock Masked autoencoders are scalable vision learners.
\newblock In {\em IEEE Conf. Comput. Vis. Pattern Recog.}, pages 16000--16009,
  2022.

\bibitem{moco}
K.~He, H.~Fan, Y.~Wu, S.~Xie, and R.~B. Girshick.
\newblock Momentum contrast for unsupervised visual representation learning.
\newblock In {\em IEEE Conf. Comput. Vis. Pattern Recog.}, pages 9726--9735,
  2020.

\bibitem{he2019rethinking}
K.~He, R.~Girshick, and P.~Doll{\'a}r.
\newblock Rethinking imagenet pre-training.
\newblock In {\em Int. Conf. Comput. Vis.}, pages 4918--4927, 2019.

\bibitem{maskrcnn}
K.~He, G.~Gkioxari, P.~Doll{\'a}r, and R.~Girshick.
\newblock Mask r-cnn.
\newblock In {\em Int. Conf. Comput. Vis.}, pages 2961--2969, 2017.

\bibitem{xlearner}
Y.~He, G.~Huang, S.~Chen, J.~Teng, W.~Kun, Z.~Yin, L.~Sheng, Z.~Liu, Y.~Qiao,
  and J.~Shao.
\newblock X-learner: Learning cross sources and tasks for universal visual
  representation.
\newblock In {\em Eur. Conf. Comput. Vis.}, 2022.

\bibitem{gelu}
D.~Hendrycks and K.~Gimpel.
\newblock Gaussian error linear units (gelus).
\newblock {\em arXiv preprint arXiv:1606.08415}, 2016.

\bibitem{inat18}
G.~V. Horn, O.~M. Aodha, Y.~Song, Y.~Cui, C.~Sun, A.~Shepard, H.~Adam,
  P.~Perona, and S.~J. Belongie.
\newblock The inaturalist species classification and detection dataset.
\newblock In {\em IEEE Conf. Comput. Vis. Pattern Recog.}, pages 8769--8778,
  2018.

\bibitem{adapter}
N.~Houlsby, A.~Giurgiu, S.~Jastrzebski, B.~Morrone, Q.~De~Laroussilhe,
  A.~Gesmundo, M.~Attariyan, and S.~Gelly.
\newblock Parameter-efficient transfer learning for nlp.
\newblock In {\em Int. Conf. Mach. Learn.}, pages 2790--2799, 2019.

\bibitem{lora}
E.~J. Hu, P.~Wallis, Z.~Allen-Zhu, Y.~Li, S.~Wang, L.~Wang, W.~Chen, et~al.
\newblock Lora: Low-rank adaptation of large language models.
\newblock In {\em Int. Conf. Learn. Represent.}, 2021.

\bibitem{gqa}
D.~A. Hudson and C.~D. Manning.
\newblock Gqa: A new dataset for real-world visual reasoning and compositional
  question answering.
\newblock In {\em IEEE Conf. Comput. Vis. Pattern Recog.}, pages 6700--6709,
  2019.

\bibitem{vpt}
M.~Jia, L.~Tang, B.-C. Chen, C.~Cardie, S.~Belongie, B.~Hariharan, and S.-N.
  Lim.
\newblock Visual prompt tuning.
\newblock In {\em Eur. Conf. Comput. Vis.}, 2022.

\bibitem{convpass}
S.~Jie and Z.-H. Deng.
\newblock Convolutional bypasses are better vision transformer adapters.
\newblock {\em arXiv preprint arXiv:2207.07039}, 2022.

\bibitem{foox251}
P.~Kaur, K.~Sikka, W.~Wang, S.~Belongie, and A.~Divakaran.
\newblock Foodx-251: a dataset for fine-grained food classification.
\newblock {\em arXiv preprint arXiv:1907.06167}, 2019.

\bibitem{stanforddogs}
A.~Khosla, N.~Jayadevaprakash, B.~Yao, and F.-F. Li.
\newblock Novel dataset for fine-grained image categorization: Stanford dogs.
\newblock In {\em Proc. CVPR workshop on fine-grained visual categorization
  (FGVC)}, 2011.

\bibitem{fpn}
A.~Kirillov, R.~B. Girshick, K.~He, and P.~Doll{\'{a}}r.
\newblock Panoptic feature pyramid networks.
\newblock In {\em IEEE Conf. Comput. Vis. Pattern Recog.}, pages 6399--6408,
  2019.

\bibitem{ubernet}
I.~Kokkinos.
\newblock Ubernet: Training a universal convolutional neural network for low-,
  mid-, and high-level vision using diverse datasets and limited memory.
\newblock In {\em IEEE Conf. Comput. Vis. Pattern Recog.}, pages 6129--6138,
  2017.

\bibitem{stanfordcars}
J.~Krause, M.~Stark, J.~Deng, and L.~Fei-Fei.
\newblock 3d object representations for fine-grained categorization.
\newblock In {\em IEEE Conf. Comput. Vis. Pattern Recog.}, pages 554--561,
  2013.

\bibitem{VG}
R.~Krishna, Y.~Zhu, O.~Groth, J.~Johnson, K.~Hata, J.~Kravitz, S.~Chen,
  Y.~Kalantidis, L.-J. Li, D.~A. Shamma, et~al.
\newblock Visual genome: Connecting language and vision using crowdsourced
  dense image annotations.
\newblock {\em Int. J. Comput. Vis.}, pages 32--73, 2017.

\bibitem{lester2021power}
B.~Lester, R.~Al-Rfou, and N.~Constant.
\newblock The power of scale for parameter-efficient prompt tuning.
\newblock {\em arXiv preprint arXiv:2104.08691}, 2021.

\bibitem{vitdet}
Y.~Li, H.~Mao, R.~Girshick, and K.~He.
\newblock Exploring plain vision transformer backbones for object detection.
\newblock {\em arXiv preprint arXiv:2203.16527}, 2022.

\bibitem{coco}
T.-Y. Lin, M.~Maire, S.~Belongie, J.~Hays, P.~Perona, D.~Ramanan,
  P.~Doll{\'a}r, and C.~L. Zitnick.
\newblock Microsoft coco: Common objects in context.
\newblock In {\em Eur. Conf. Comput. Vis.}, pages 740--755, 2014.

\bibitem{prompt_survery}
P.~Liu, W.~Yuan, J.~Fu, Z.~Jiang, H.~Hayashi, and G.~Neubig.
\newblock Pre-train, prompt, and predict: A systematic survey of prompting
  methods in natural language processing.
\newblock {\em arXiv preprint arXiv:2107.13586}, 2021.

\bibitem{ptuning-v2}
X.~Liu, K.~Ji, Y.~Fu, Z.~Du, Z.~Yang, and J.~Tang.
\newblock P-tuning v2: Prompt tuning can be comparable to fine-tuning
  universally across scales and tasks.
\newblock {\em arXiv preprint arXiv:2110.07602}, 2021.

\bibitem{liu2021gpt}
X.~Liu, Y.~Zheng, Z.~Du, M.~Ding, Y.~Qian, Z.~Yang, and J.~Tang.
\newblock Gpt understands, too.
\newblock {\em arXiv preprint arXiv:2103.10385}, 2021.

\bibitem{place365}
A.~L{\'{o}}pez{-}Cifuentes, M.~Escudero{-}Vi{\~{n}}olo, J.~Besc{\'{o}}s, and
  {\'{A}}.~Garc{\'{\i}}a{-}Mart{\'{\i}}n.
\newblock Semantic-aware scene recognition.
\newblock {\em Pattern Recognit.}, page 107256, 2020.

\bibitem{unified-io}
J.~Lu, C.~Clark, R.~Zellers, R.~Mottaghi, and A.~Kembhavi.
\newblock Unified-io: A unified model for vision, language, and multi-modal
  tasks.
\newblock {\em arXiv preprint arXiv:2206.08916}, 2022.

\bibitem{flowers}
M.-E. Nilsback and A.~Zisserman.
\newblock Automated flower classification over a large number of classes.
\newblock In {\em 2008 Sixth Indian Conference on Computer Vision, Graphics \&
  Image Processing}, pages 722--729, 2008.

\bibitem{stilts}
J.~Phang, T.~F{\'e}vry, and S.~R. Bowman.
\newblock Sentence encoders on stilts: Supplementary training on intermediate
  labeled-data tasks.
\newblock {\em arXiv preprint arXiv:1811.01088}, 2018.

\bibitem{DBLP:conf/emnlp/PothPRG21}
C.~Poth, J.~Pfeiffer, A.~R{\"{u}}ckl{\'{e}}, and I.~Gurevych.
\newblock What to pre-train on? efficient intermediate task selection.
\newblock In {\em Annual Conference on Empirical Methods in Natural Language
  Processing}, pages 10585--10605, 2021.

\bibitem{pruksachatkun2020intermediate}
Y.~Pruksachatkun, J.~Phang, H.~Liu, P.~M. Htut, X.~Zhang, R.~Y. Pang, C.~Vania,
  K.~Kann, and S.~R. Bowman.
\newblock Intermediate-task transfer learning with pretrained models for
  natural language understanding: When and why does it work?
\newblock {\em arXiv preprint arXiv:2005.00628}, 2020.

\bibitem{clip}
A.~Radford, J.~W. Kim, C.~Hallacy, A.~Ramesh, G.~Goh, S.~Agarwal, G.~Sastry,
  A.~Askell, P.~Mishkin, J.~Clark, et~al.
\newblock Learning transferable visual models from natural language
  supervision.
\newblock In {\em Int. Conf. Mach. Learn.}, pages 8748--8763, 2021.

\bibitem{in21k}
T.~Ridnik, E.~Ben-Baruch, A.~Noy, and L.~Zelnik-Manor.
\newblock Imagenet-21k pretraining for the masses.
\newblock {\em arXiv preprint arXiv:2104.10972}, 2021.

\bibitem{ruder2017overview}
S.~Ruder.
\newblock An overview of multi-task learning in deep neural networks.
\newblock {\em arXiv preprint arXiv:1706.05098}, 2017.

\bibitem{overfeat}
P.~Sermanet, D.~Eigen, X.~Zhang, M.~Mathieu, R.~Fergus, and Y.~LeCun.
\newblock Overfeat: Integrated recognition, localization and detection using
  convolutional networks.
\newblock In {\em Int. Conf. Learn. Represent.}, 2014.

\bibitem{intern}
J.~Shao, S.~Chen, Y.~Li, K.~Wang, Z.~Yin, Y.~He, J.~Teng, Q.~Sun, M.~Gao,
  J.~Liu, et~al.
\newblock Intern: A new learning paradigm towards general vision.
\newblock {\em arXiv preprint arXiv:2111.08687}, 2021.

\bibitem{objects365}
S.~Shao, Z.~Li, T.~Zhang, C.~Peng, G.~Yu, X.~Zhang, J.~Li, and J.~Sun.
\newblock Objects365: A large-scale, high-quality dataset for object detection.
\newblock In {\em Int. Conf. Comput. Vis.}, pages 8430--8439, 2019.

\bibitem{moe}
N.~Shazeer, A.~Mirhoseini, K.~Maziarz, A.~Davis, Q.~Le, G.~Hinton, and J.~Dean.
\newblock Outrageously large neural networks: The sparsely-gated
  mixture-of-experts layer.
\newblock {\em arXiv preprint arXiv:1701.06538}, 2017.

\bibitem{zoo_tuning}
Y.~Shu, Z.~Kou, Z.~Cao, J.~Wang, and M.~Long.
\newblock Zoo-tuning: Adaptive transfer from a zoo of models.
\newblock In {\em Int. Conf. Mach. Learn.}, pages 9626--9637, 2021.

\bibitem{nyudepth}
N.~Silberman, D.~Hoiem, P.~Kohli, and R.~Fergus.
\newblock Indoor segmentation and support inference from rgbd images.
\newblock {\em Eur. Conf. Comput. Vis.}, pages 746--760, 2012.

\bibitem{jft}
C.~Sun, A.~Shrivastava, S.~Singh, and A.~Gupta.
\newblock Revisiting unreasonable effectiveness of data in deep learning era.
\newblock In {\em Int. Conf. Comput. Vis.}, pages 843--852, 2017.

\bibitem{tan2018survey}
C.~Tan, F.~Sun, T.~Kong, W.~Zhang, C.~Yang, and C.~Liu.
\newblock A survey on deep transfer learning.
\newblock In {\em International conference on artificial neural networks},
  pages 270--279, 2018.

\bibitem{nabirds}
G.~Van~Horn, S.~Branson, R.~Farrell, S.~Haber, J.~Barry, P.~Ipeirotis,
  P.~Perona, and S.~Belongie.
\newblock Building a bird recognition app and large scale dataset with citizen
  scientists: The fine print in fine-grained dataset collection.
\newblock In {\em IEEE Conf. Comput. Vis. Pattern Recog.}, pages 595--604,
  2015.

\bibitem{mtinet}
S.~Vandenhende, S.~Georgoulis, and L.~V. Gool.
\newblock Mti-net: Multi-scale task interaction networks for multi-task
  learning.
\newblock In {\em Eur. Conf. Comput. Vis.}, pages 527--543, 2020.

\bibitem{transformer}
A.~Vaswani, N.~Shazeer, N.~Parmar, J.~Uszkoreit, L.~Jones, A.~N. Gomez,
  {\L}.~Kaiser, and I.~Polosukhin.
\newblock Attention is all you need.
\newblock {\em Adv. Neural Inform. Process. Syst.}, 2017.

\bibitem{cub200}
C.~Wah, S.~Branson, P.~Welinder, P.~Perona, and S.~Belongie.
\newblock The caltech-ucsd birds-200-2011 dataset.
\newblock 2011.

\bibitem{logodet}
J.~Wang, W.~Min, S.~Hou, S.~Ma, Y.~Zheng, and S.~Jiang.
\newblock Logodet-3k: A large-scale image dataset for logo detection.
\newblock {\em ACM Transactions on Multimedia Computing, Communications, and
  Applications (TOMM)}, pages 1--19, 2022.

\bibitem{logo2k}
J.~Wang, W.~Min, S.~Hou, S.~Ma, Y.~Zheng, H.~Wang, and S.~Jiang.
\newblock Logo-2k+: A large-scale logo dataset for scalable logo
  classification.
\newblock In {\em Assoc. Adv. Artif. Intell.}, pages 6194--6201, 2020.

\bibitem{loveda}
J.~Wang, Z.~Zheng, A.~Ma, X.~Lu, and Y.~Zhong.
\newblock Loveda: {A} remote sensing land-cover dataset for domain adaptive
  semantic segmentation.
\newblock In J.~Vanschoren and S.~Yeung, editors, {\em Proceedings of the
  Neural Information Processing Systems Track on Datasets and Benchmarks 1,
  NeurIPS Datasets and Benchmarks 2021}, 2021.

\bibitem{ofa}
P.~Wang, A.~Yang, R.~Men, J.~Lin, S.~Bai, Z.~Li, J.~Ma, C.~Zhou, J.~Zhou, and
  H.~Yang.
\newblock Ofa: Unifying architectures, tasks, and modalities through a simple
  sequence-to-sequence learning framework.
\newblock In {\em Int. Conf. Mach. Learn.}, pages 23318--23340, 2022.

\bibitem{beit-3}
W.~Wang, H.~Bao, L.~Dong, J.~Bjorck, Z.~Peng, Q.~Liu, K.~Aggarwal, O.~K.
  Mohammed, S.~Singhal, S.~Som, et~al.
\newblock Image as a foreign language: Beit pretraining for all vision and
  vision-language tasks.
\newblock {\em arXiv preprint arXiv:2208.10442}, 2022.

\bibitem{fewshot_survey}
Y.~Wang, Q.~Yao, J.~T. Kwok, and L.~M. Ni.
\newblock Generalizing from a few examples: A survey on few-shot learning.
\newblock {\em ACM computing surveys (csur)}, 53(3):1--34, 2020.

\bibitem{upernet}
T.~Xiao, Y.~Liu, B.~Zhou, Y.~Jiang, and J.~Sun.
\newblock Unified perceptual parsing for scene understanding.
\newblock In {\em Eur. Conf. Comput. Vis.}, pages 432--448, 2018.

\bibitem{simmim}
Z.~Xie, Z.~Zhang, Y.~Cao, Y.~Lin, J.~Bao, Z.~Yao, Q.~Dai, and H.~Hu.
\newblock Simmim: A simple framework for masked image modeling.
\newblock In {\em IEEE Conf. Comput. Vis. Pattern Recog.}, pages 9653--9663,
  2022.

\bibitem{padnet}
D.~Xu, W.~Ouyang, X.~Wang, and N.~Sebe.
\newblock Pad-net: Multi-tasks guided prediction-and-distillation network for
  simultaneous depth estimation and scene parsing.
\newblock In {\em IEEE Conf. Comput. Vis. Pattern Recog.}, pages 675--684,
  2018.

\bibitem{pevl}
Y.~Yao, Q.~Chen, A.~Zhang, W.~Ji, Z.~Liu, T.-S. Chua, and M.~Sun.
\newblock Pevl: Position-enhanced pre-training and prompt tuning for
  vision-language models.
\newblock In {\em Annual Conference on Empirical Methods in Natural Language
  Processing}, 2022.

\bibitem{Flickr}
P.~Young, A.~Lai, M.~Hodosh, and J.~Hockenmaier.
\newblock From image descriptions to visual denotations: New similarity metrics
  for semantic inference over event descriptions.
\newblock {\em Transactions of the Association for Computational Linguistics},
  pages 67--78, 2014.

\bibitem{coca}
J.~Yu, Z.~Wang, V.~Vasudevan, L.~Yeung, M.~Seyedhosseini, and Y.~Wu.
\newblock Coca: Contrastive captioners are image-text foundation models.
\newblock {\em arXiv preprint arXiv:2205.01917}, 2022.

\bibitem{refcoco}
L.~Yu, P.~Poirson, S.~Yang, A.~C. Berg, and T.~L. Berg.
\newblock Modeling context in referring expressions.
\newblock In {\em Eur. Conf. Comput. Vis.}, pages 69--85, 2016.

\bibitem{yu2020gradient}
T.~Yu, S.~Kumar, A.~Gupta, S.~Levine, K.~Hausman, and C.~Finn.
\newblock Gradient surgery for multi-task learning.
\newblock {\em Adv. Neural Inform. Process. Syst.}, pages 5824--5836, 2020.

\bibitem{taskonomy}
A.~R. Zamir, A.~Sax, W.~Shen, L.~J. Guibas, J.~Malik, and S.~Savarese.
\newblock Taskonomy: Disentangling task transfer learning.
\newblock In {\em IEEE Conf. Comput. Vis. Pattern Recog.}, pages 3712--3722,
  2018.

\bibitem{vcr}
R.~Zellers, Y.~Bisk, A.~Farhadi, and Y.~Choi.
\newblock From recognition to cognition: Visual commonsense reasoning.
\newblock In {\em IEEE Conf. Comput. Vis. Pattern Recog.}, pages 6720--6731,
  2019.

\bibitem{vtab}
X.~Zhai, J.~Puigcerver, A.~Kolesnikov, P.~Ruyssen, C.~Riquelme, M.~Lucic,
  J.~Djolonga, A.~S. Pinto, M.~Neumann, A.~Dosovitskiy, et~al.
\newblock A large-scale study of representation learning with the visual task
  adaptation benchmark.
\newblock {\em arXiv preprint arXiv:1910.04867}, 2019.

\bibitem{tip-adapter}
R.~Zhang, R.~Fang, P.~Gao, W.~Zhang, K.~Li, J.~Dai, Y.~Qiao, and H.~Li.
\newblock Tip-adapter: Training-free clip-adapter for better vision-language
  modeling.
\newblock {\em arXiv preprint arXiv:2111.03930}, 2021.

\bibitem{noah}
Y.~Zhang, K.~Zhou, and Z.~Liu.
\newblock Neural prompt search.
\newblock {\em arXiv preprint arXiv:2206.04673}, 2022.

\bibitem{jtrl}
Z.~Zhang, Z.~Cui, C.~Xu, Z.~Jie, X.~Li, and J.~Yang.
\newblock Joint task-recursive learning for semantic segmentation and depth
  estimation.
\newblock In {\em Eur. Conf. Comput. Vis.}, pages 235--251, 2018.

\bibitem{ade20k}
B.~Zhou, H.~Zhao, X.~Puig, S.~Fidler, A.~Barriuso, and A.~Torralba.
\newblock Scene parsing through {ADE20K} dataset.
\newblock In {\em IEEE Conf. Comput. Vis. Pattern Recog.}, pages 5122--5130,
  2017.

\bibitem{thdogs}
D.-N. Zou, S.-H. Zhang, T.-J. Mu, and M.~Zhang.
\newblock A new dataset of dog breed images and a benchmark for finegrained
  classification.
\newblock {\em Computational Visual Media}, pages 477--487, 2020.

\end{thebibliography}
}


\end{document}